\documentclass[runningheads]{llncs}
\usepackage[T1]{fontenc}
\usepackage{graphicx}
\usepackage{amsmath, amsfonts}
\usepackage{mathrsfs}
\newcommand{\R}{\mathbb{R}}

\begin{document}
\title{A Comparative Study on Energy Consumption Models for Drones}
%
%
\author{Carlos~Muli\inst{1} \and
Sangyoung~Park\inst{2} \and
Mingming~Liu\inst{1}}
\authorrunning{C. Muli et al.}
%
\institute{School of Electronic Engineering, Dublin City University, Dublin, Ireland \and Einstein Center Digital Future, Technical University of Berlin, Berlin, Germany \\
\email{carlos.muli2@mail.dcu.ie; sangyoung.park@tu-berlin.de; mingming.liu@dcu.ie}}
\maketitle              
\begin{abstract}
Creating an appropriate energy consumption prediction model is becoming an important topic for drone-related research in the literature. However, a general consensus on the energy consumption model is yet to be reached at present. As a result, there are many variations that attempt to create models that range in complexity with a focus on different aspects. In this paper, we benchmark the five most popular energy consumption models for drones derived from their physical behaviours and point to the difficulties in matching with a realistic energy dataset collected from a delivery drone in flight under different testing conditions. Moreover, we propose a novel data-driven energy model using the Long Short-Term Memory (LSTM) based deep learning architecture and the accuracy is compared based on the dataset. Our experimental results have shown that the LSTM based approach can easily outperform other mathematical models for the dataset under study. Finally, sensitivity analysis has been carried out in order to interpret the model. 

\end{abstract}

\section{Introduction}

Unmanned Aerial Vehicles (UAVs) are being used in many diverse operations in a range of fields at present, including but not limited to aerial surveillance, search and rescue operations, parcel delivery, and agriculture \cite{droneUse,zaheer2016aerial}. However, one of the most critical design issues for UAVs is that they often suffer from short flight time, typically tens of minutes, mainly due to their high power requirements and limited battery capacity \cite{thibbotuwawa2018energy,hu2020energy}. Thus, it is important to correctly estimate the flight time and range to ensure reliable operation and energy-efficient path planning. To accomplish this, an accurate drone energy consumption model is essential, which enables quantifying the impact of different factors, i.e., wind speed, payload, ground speed, altitude, etc., affecting the energy consumption of drones in various scenarios.

However, it is a challenging task to incorporate all such factors into a single energy consumption model. There are numerous factors related to the drone design (weight, number of rotors, battery weight/efficiency, avionics), the environment (wind conditions, weather, ambient temperature), drone dynamics (acceleration, angle of attack, flight angle/altitude), and operational requirements (flight time, payload). Existing models on drone energy consumption focuses on certain aspects of drones and usually consider a subset of factors~\cite{ref1}. Some models can only be applied to particular conditions such as hovering largely ignoring the impact of the air speed~\cite{model2}. There are models focusing more on drones acting like fixed-wing aircrafts~\cite{ref4} or more like helicopters~\cite{model2}. This results in conflicting predictions across different energy consumption models despite same input parameters and trajectories~\cite{ref1}. Apart from the models taking theoretical approaches~\cite{ref4,model2,model3,model4}, there have been attempts to use regression models~\cite{model5,Prasetia} to match the model better to the realistic measurements. In particular, black box modeling of drones also reports decent results across the missions starting from take-off to return~\cite{Prasetia}. The work also reports that to account for impacts of control profiles, which is often ignored in other models, a time-series machine learning methods needs to be investigated.

In this paper, we aim at comparing some prominent drone energy consumption models in the literature and propose a long-short term memory (LSTM) deep learning-based architecture, useful for prediction of time-series data, for energy consumption model of drones. Despite such efforts to accurately model the drone energy consumption and decent accuracy reported in the literature, there has been few works on directly comparing different types of models on a real measured data. Our efforts to apply the energy consumption models to a recently published measurement data~\cite{ref5}, have shown that significant discrepancies exist between the predicted values and the independently-collected real-world measurement. Therefore, we consider a deep learning approach to create an energy consumption model that considers all aspects of flight (take-off, landing, cruising, hovering) from empirical data and the prediction results have been compared with model-based approaches fitted to the real-world data. 

The key contributions in this paper can be summarized as follows:

\begin{itemize}
	\item We evaluate the performance of several existing drone energy consumption models using the specific realistic dataset \cite{ref5}.
	\item We propose a learning-based approach using the Long-Short Term Memory (LSTM) deep learning based architecture for power consumption prediction of drones. 
	\item We carry out sensitivity analysis on the trained LSTM model to give some insights on feature importance, i.e., interpretability of the model.
\end{itemize}

The remainder of this paper is organised as follows. Section \ref{TB} elaborates the technical background on some prominent drone energy consumption models in the literature. Section \ref{design} presents our key research problem and propose the LSTM-based system architecture to address the design issue. Section \ref{result} discusses our evaluation results using different models. Finally, Section \ref{conclusion} concludes the paper.

\section{Background: Drone Power Models} \label{TB}
In this section, we introduce drone energy models used in the literature.
Prior works on path planning algorithms assume drone energy models for evaluating battery usage.
As the flight distance and the number of turns a drone makes determines the paths, energy consumption models regards the factors important in estimating the energy consumption~\cite{ref2}.

Other factors such as wind speed, wind angle, and altitude are not considered in the work.
More elaborate models consider the impact of acceleration and deceleration~\cite{ref3}.
Several tests were performed in this study which focuses on three main performance metrics: straight line distance, the effects of velocity, and the effects of turning.
The study has shown that higher speeds result in lower overall energy consumption and higher turning angles resulted in higher overall energy consumption.
Analysis of all aspects of on-board electronics to form total energy consumption is also explored and is validated using empirical data from a commercial drone~\cite{additionalref1}. Energy consumption primarily comes from the motors, followed by communications, processors, and sensors. Communications, processors, and sensors were discovered to be minimal in contribution but not negligible in total energy consumption.

Among such models, we selected representative ones to be investigated in this paper also designated as ``five fundamental models for drone energy consumption of steady level flight’' in a recent survey~\cite{ref1}, dubbed D’Andrea, Dorling et al., Stolaroff et al., Kirchstein, and Tseng energy models.
\subsection{D’Andrea Energy Model}
The D’Andrea energy model is based on the drone’s lift-to-drag ratio~\cite{model1,ref1}. The formula is optimised for steady drone flight and uses the drone’s mass, airspeed, lift-to-drag ratio, and the power transfer efficiency of the battery. The model comes in two variations; a standard variation with no account for wind, and one that does account for wind in terms of headwind experienced by the drone.
The model is further expanded by implementing “empty returns”, which occurs when the drone drops off the payload before taking the return flight~\cite{ref4}.
The model makes several assumptions to create their energy consumption model. The payload of the drone is no heavier than 2~kg and has an operating range of 10~km. A lift-to-drag ratio of a constant value is selected, inspired by helicopter lift-to-drag ratios and used for comparison with them. Variables such as cruising speed during missions are set to a predetermined value of 45 km/h. The power transfer efficiency is set to 0.5 \cite{model1}. A constant $p_{avio}$ is added to account for vehicle avionics.
\begin{equation}
P = \frac{\sum_{k=1}^{3} m_k v_a}{370 \eta r} + p_{avio},
\end{equation}
where $m_k$ represents the mass of each drone component including drone weight $(k=1)$, battery weight $(k=2)$ and payload weight $(k=3)$. $v_a$ is the drone airspeed, i.e., the speed of drone relative to air, $\eta$ is the power transfer efficiency, $r$ is the lift-to-drag ratio, and $p_{avio}$ is the power required for drone avionics \cite{ref1}.
\subsection{Dorling et al. Energy Model}
The Dorling energy model only takes into consideration drone hovering, and thus, cannot detail energy consumption for the take-off, cruising, and landing~ \cite{model2,ref1}. However, this model does consider the components used in the drone such as the number of rotors and propeller area. Through testing, the equations that dictate the energy consumption was reduced to a linear function dependent on the battery and payload of the drone.
The model is derived from the equation used to calculate the power of a helicopter and is adapted for multi-rotors. The mass components of drone $m_k$ are used as parameters alongside gravity ($g$), air density ($\rho$), number of rotors ($n$), and propeller area ($\zeta$).
\begin{equation}
P = \frac{g (\sum_{k=1}^{3} m_k) ^{\frac{3}{2}}}{\sqrt{2n\rho\zeta}}
\end{equation}
\subsection{Stolaroff et al. Energy Model}
The Stolaroff energy model designs its model using the physics of drone flight including the forces experienced by the drone due to its weight, parasitic drag, and induced drag~\cite{model3,ref1}.
The model accounts for heavy winds by utilising an adapted version of the previous model by using the angle of attack of the drone. However, it was noted that large values of the angle of attack resulted in unstable results. The model consists of the thrust produced ($T$), angle of attack ($\alpha$), power transfer efficiency ($\eta$), and the induced speed caused by the drone ($v_i$). It can be presented as follows:
\begin{equation}
P = \frac{T ( v_a sin(\alpha) + v_i)} {\eta}
\end{equation}
where $T=g\sum_{k=1}^{3}m_k + 0.5 \rho \sum_{k=1}^{3} C_{D_k} A_k v_a^{2}$ with the drag coefficient $C_{D_k}$, and the projected area perpendicular to travel of each drone component  $A_k$ \cite{ref1}. 
\subsection{Kirchstein Energy Model}
The Kirchstein energy model is based on the drone’s environmental conditions and flight trajectory~\cite{model4,ref1}. It is another component model with a focus on optimised take-off angle, cruising altitude, level flight, descent, and landing. This model takes into consideration a wide range of factors such as the power required for climbing, avionics, and different power losses resulting from the electric motor and power transmission inefficiencies. The model covers the power consumption from air drag from the drone’s profile and the rotor profile, the lift required for flight, the climb to the designated altitude, and power supplied to any electronics on-board.
\begin{equation}
	P = \frac{1}{\eta} (\kappa Tw + \frac{1}{2}\rho(\sum_{k=1}^{3} C_{D_k} A_k)v_a^{3} + \kappa_2(g\sum_{k=1}^{3} m_k)^{1.5} +\kappa_3(g\sum_{k=1}^{3} m_k)^{0.5} v_a^{2}) + \frac{P_{avio}}{\eta_c},
\end{equation}
where $\kappa, \kappa_2, \kappa_3$ are constants, $w$ is the downwash coefficient, and $\eta_c$ is the battery charging efficiency \cite{ref1}.
\subsection{Tseng Energy Model}
The Tseng energy model differs from the other energy models as it consists of a nine-term nonlinear regression model created from collected data~\cite{model5}. This model was created from horizontal and vertical speeds and accelerations, payload, mass, and wind speed data gathered from empirical testing. The drone used for data collection was a DJI Matrice 100 and tested for payloads of 0, 0.3, and 0.6 kg. The drone used for testing consisted of three experiments that recorded the drone’s ability to hover without any input or movement, ability to climb and descend, and ability to move horizontally. The model assesses the impact of motion through these tests and the payload weights to create the model. A 3DR Solo drone was also used to create an alternative version of the model for smaller sized drones with the expression for energy consumption shown below. Note that the model is essentially a function of payload mass $m_3$ and the airspeed $v_a$ \cite{ref1}.
\begin{equation}
P = -2.595v_a + 0.197 m_3 + 251.7.
\end{equation}

\section{Data-Driven Model and Proposed LSTM Architecture} \label{design}

In this section, we introduce the data-driven model to fit the realistic dataset~\cite{ref5}.  We first present the problem statement for the model fitting task, then we present details for the dataset, and finally we demonstrate the proposed architecture.

\subsection{Problem Statement}
The key problem that we are considering here is to build an energy consumption model which can take a sequence of input feature data from a drone and predict the corresponding energy consumption of the drone as output. More specifically, the input data essentially captures the characteristics, dynamics and environmental context, e.g., wind speed, payload, ground speed, for the drone under test in a given scenario. 

Mathematically, let $F_{t} \in \R^{N}$ be the feature vector consisting of the $N$ features for the drone at time $t$. Let $E_{t} \in \R$ denote the energy consumption of the drone at time $t$. For a given time window $\mathcal{T}: =\left\lbrace 1 , 2, \dots, T \right\rbrace$, our objective is to find a learning function $H(.)$ which is able to address the following problem:

\begin{equation}\label{localopt}
\begin{aligned}
\min_{H} \quad & \sum_{t \in \mathcal{T}}  (E_{t} - \hat{E}_{t})^2 \\
\textrm{s.t.} \quad & \hat{E}_{t} =  H(F_{t}) \\
\end{aligned}
\end{equation}
where $\hat{E}_{t}$ denotes the predicted energy consumption of the drone at time $t$ with respect to the input feature vector $F_{t}$.

\subsection{Experimental Dataset}
The experimental dataset used in this work is from the paper \cite{ref5}. The dataset presents some very recent energy consumption information for a DJI Matrice 100 drone that consists of a total of 195 test flights with variations in payload, speed, and altitude. A total number of 28 features were recorded onboard which were taken from the battery state, Global Positioning System (GPS), Inertial Measurement Unit (IMU), and the wind measurement unit. A total number of 21 features was included from the dataset in order to create our energy prediction model, the details of which are summarized as follows:

\begin{itemize}
	\item Wind Speed: Speed of wind recorded by the anemometer in meters per second (m/s).
	\item Wind Angle: Angle of the airspeed recorded by the anemometer with respect to north in (degrees).
	\item Position X, Y, Z: Longitude, latitude and altitude recorded by the GPS (degrees).
	\item Orientation X, Y, Z, W: Orientation as recorded by the IMU in (quarternions).
	\item Velocity X, Y, Z: Ground speed recorded by the GPS and IMU in meters per second (m/s).
	\item Angular X, Y, Z: Angular velocity recorded by the IMU in radians per second (rad/s).
	\item Linear Acceleration X, Y, Z: Linear acceleration recorded by the IMU  in meters per second squared (m/s$^{2}$).
	\item Speed: Input ground speed before flight in meters per second (m/s).
	\item Payload: Payload mass attached to the drone prior to a test in grams (g).
	\item Altitude: Input altitude the drone rises to before following flight route in meters (m).
\end{itemize}

\subsection{System Architecture}
To find out the learning function $H$, we propose the LSTM architecture which is shown in the Fig. \ref{architecture} below. Specifically, the proposed LSTM architecture consists of two bidirectional LSTM layers stacked together with a dropout layer attached to the second LSTM layer before connecting to a dense layer for output. Some details for model and network setup are reported as follows: 

\begin{itemize}
	\item  The activation function used for the output layer was a tangent function.  
	\item  The number of hidden cells for each LSTM layer was defined as 128.
	\item  The length of the input time window $T$ was set as 10. 
	\item  The Adam optimizer was chosen for model training. 
	\item  The proposed model was assembled using Keras at the backend. 
\end{itemize}

In order to achieve the optimal performance for energy prediction using the proposed architecture, we considered dropout, batch size and learning rate as hyperparameters for model tuning through grid search. For each setting of the hyperparameters, a 5-fold cross-validation was carried out on the dataset to evaluate the performance of the resulting model. The best set of parameters which results in the minimum averaged mean square error (MAE) was chosen as the optimal hyperparameters for the proposed model. Table \ref{tab1} presents our results for the hyperparamters tuning in different settings. We also present the calculation results for the root mean square error (RMSE) in the table for reference, and it can be seen that the optimal hyperparameters for the model are presented in the third row of the table, and this finding is consistent with both performance metrics, i.e., averaged RMSE and averaged MAE. Finally, the training and validation loss under the optimal configuration are shown in Fig. \ref{trainingCurve}, where two curves are converging gradually in a few number of epochs. 

\begin{figure*}[htbp]
	\centering
	\includegraphics[width=1\columnwidth, height=5in]{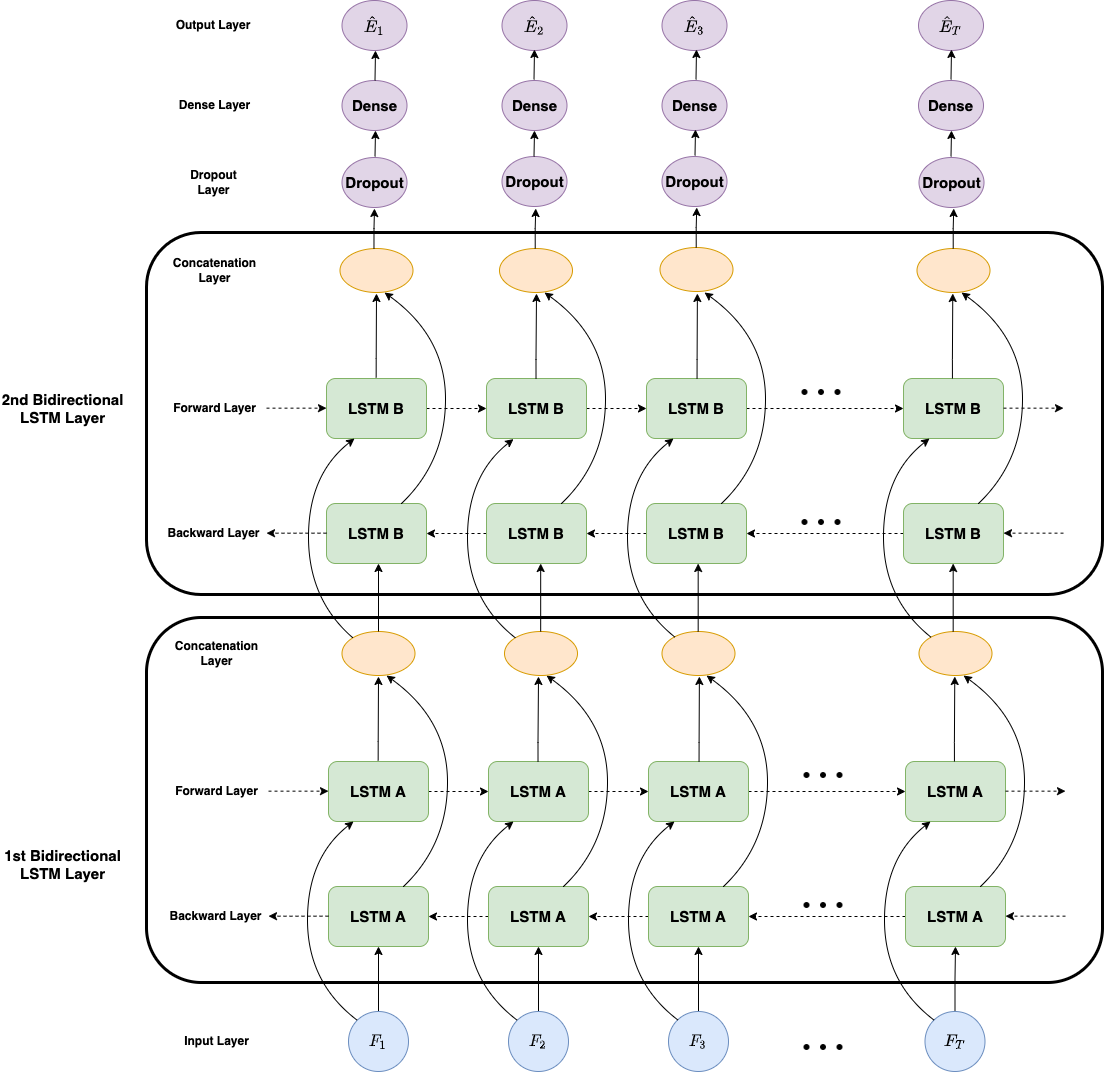}
	\caption{The proposed stacked bidirectional LSTM-based system architecture.}\label{architecture}
\end{figure*}

\begin{table}[htbp]
	\caption{Hyperparameter tuning for the proposed model}\label{tab1}
	\centering
	\begin{tabular}{|c|c|c|c|c|}
		\hline
		\textbf{Dropout} & \textbf{Learning Rate} & \textbf{Batch Size} & \textbf{\begin{tabular}[c]{@{}l@{}}Avg. \\ RMSE\end{tabular}} & \textbf{\begin{tabular}[c]{@{}l@{}}Avg. \\ MAE\end{tabular}} \\\hline
		0.2     & 0.001       & 128        & 47.4891                                              & 5.7213                                             \\ \hline
		0.2     & 0.0001    & 128        & 50.7283                                              & 6.0008                                              \\ \hline
		0.5     & 0.001       & 128        & 46.9918                                              & 5.6942                                              \\\hline
		0.5     & 0.0001    & 128        & 48.3828                                              & 5.8538                                              \\\hline
		0.5     & 0.01         & 128        & 63.1021                                              & 6.8779                                              \\\hline
		0.5     & 0.001      & 64         & 51.3855                                              & 5.9844                                              \\\hline
	\end{tabular}
\end{table}

\begin{figure}[htbp]
	\centering
	\includegraphics[width=\columnwidth, height=2.5in]{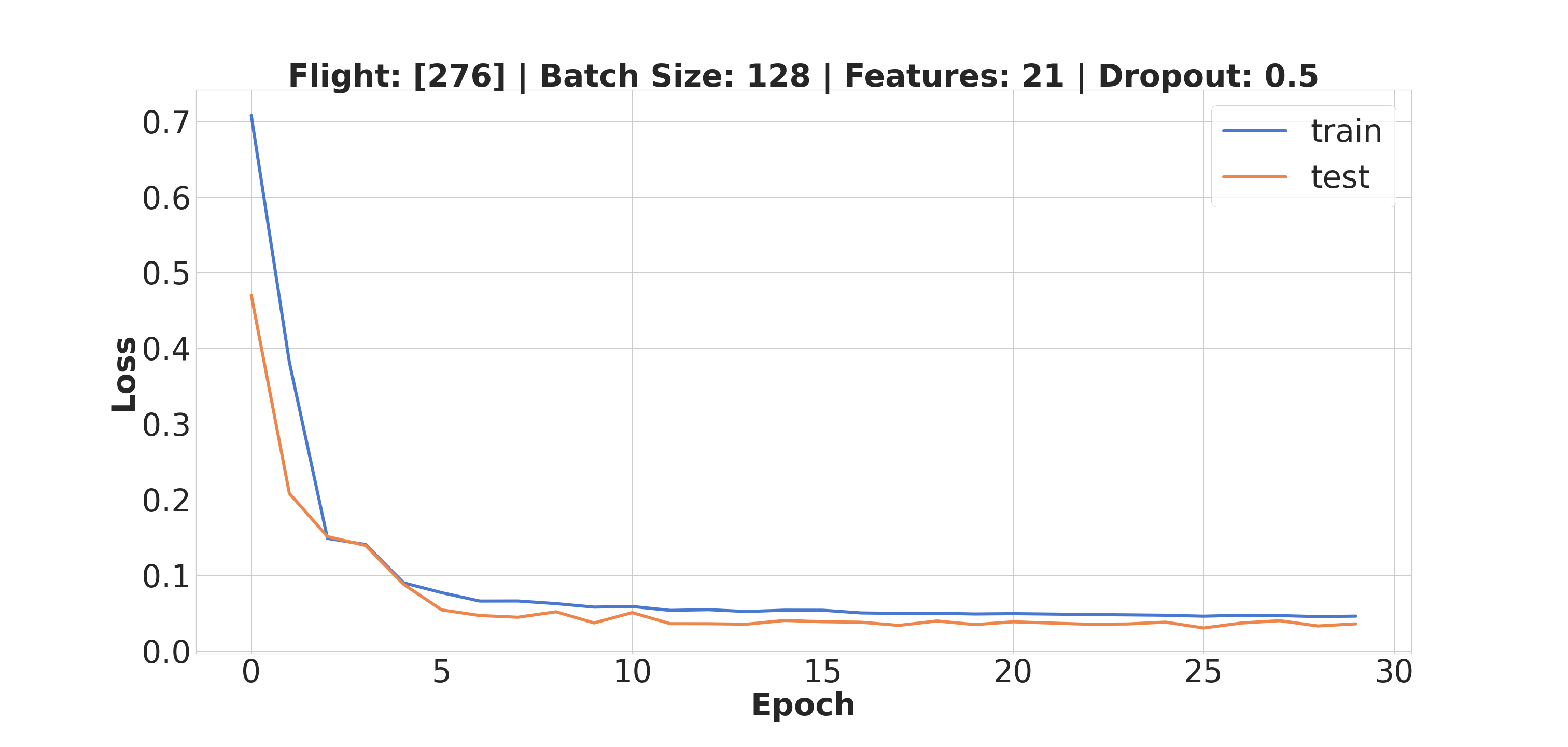}
	\caption{Training and validation loss curves using the optimal configuration.}\label{trainingCurve}
\end{figure}

\section{Results and Discussion} \label{result}

\subsection{Performance Evaluation}
\begin{figure}[htbp]
	\centering
	\includegraphics[width=\columnwidth, height=2.3in]{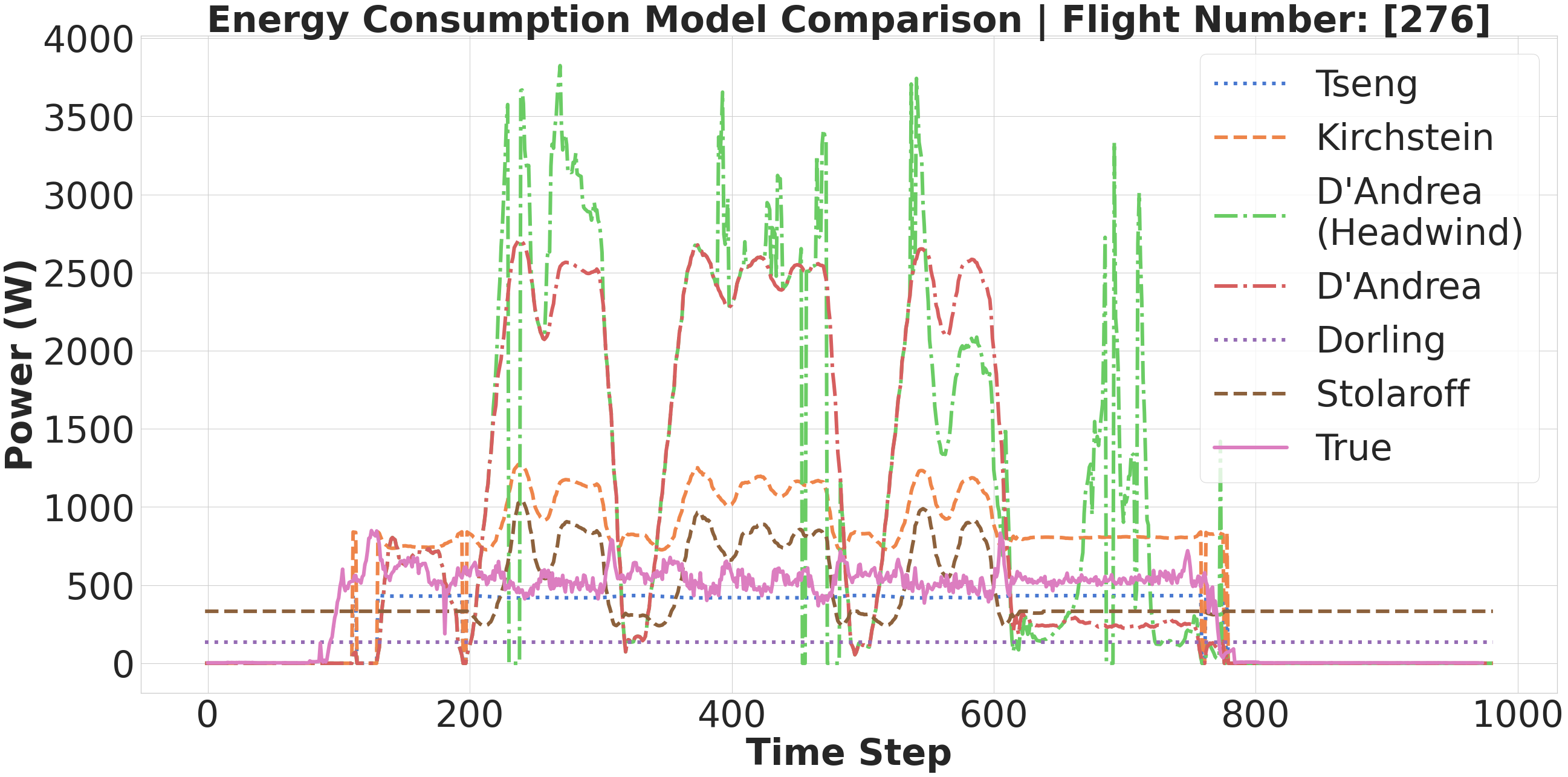}
	\caption{Comparison of the five mathematical models for energy consumption.}\label{fig15}
\end{figure}

\begin{figure}[htbp]
	\centering
	\includegraphics[width=\columnwidth, height=2.3in]{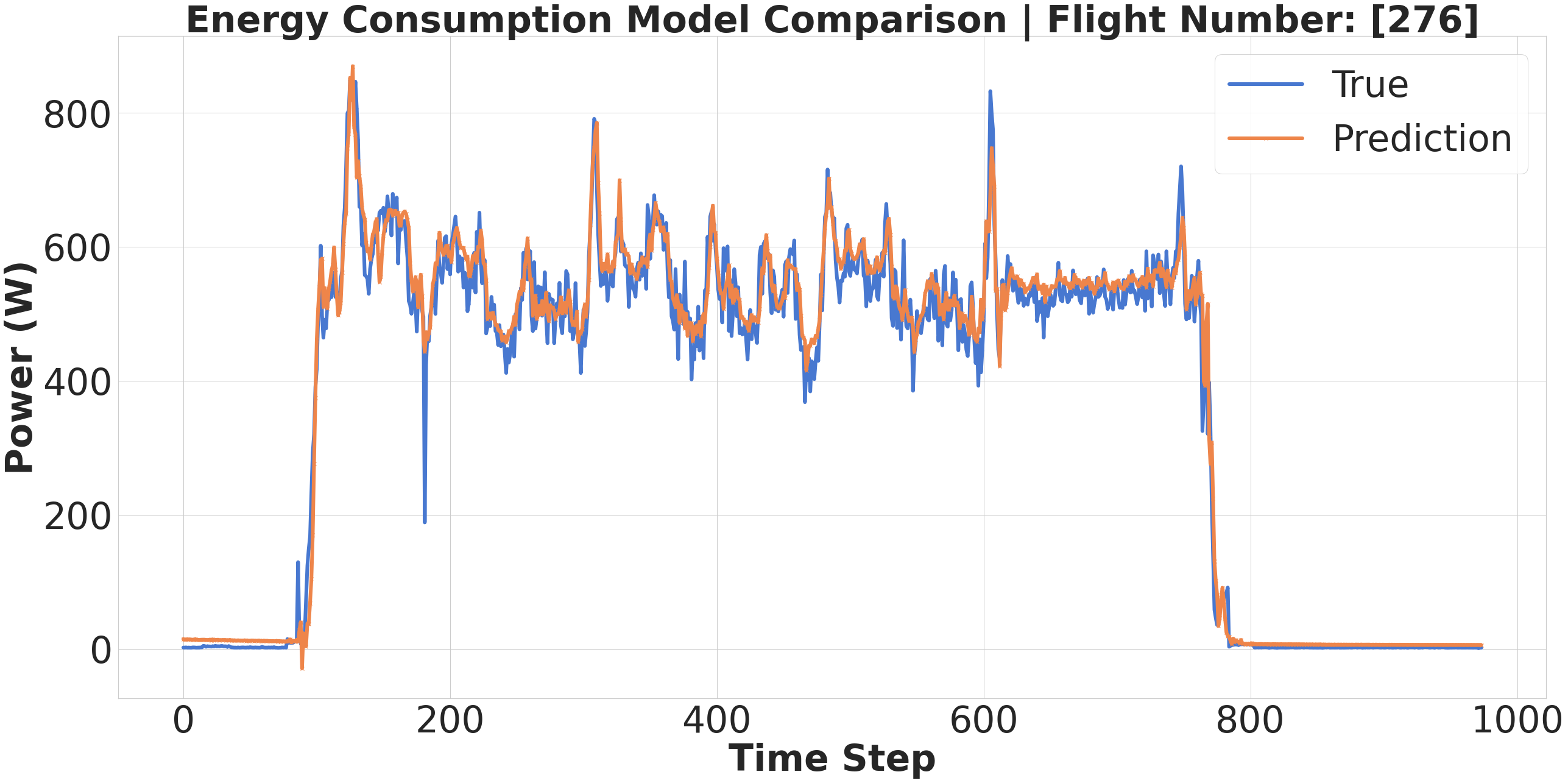}
	\caption{Comparison of the proposed LSTM model with the ground truth.}\label{lstm_result}
\end{figure}

In this section, we present our experimental results using the five prominent mathematical models as well as the proposed LSTM model in its optimal configuration. Our objective is to illustrate the differences of model prediction performance based on the realistic dataset \cite{ref5}, and reveal the superiority of the model performance using our proposed LSTM-based architecture. Specifically, we applied all models to a specific flight (flight 276) which had not been used for model training. While the five models originated from different studies, unified notation between parameters was used for fair comparison similar to the approach used in \cite{ref1}. Input parameters from the dataset were airspeed and the weight components of the drone, which were adapted to use in all models. The prediction results for different models under test are illustrated in Fig. \ref{fig15} and Fig. \ref{lstm_result}, where Fig. \ref{fig15} compared the performance for all mathematical models and Fig. \ref{lstm_result} compares performance for the LSTM-based model only. The key performance metrics are also summarised and reported in Table \ref{tab2} for ease of comparison. 

\begin{table}[htbp]
	\caption{Energy Model Comparison}\label{tab2}
	\centering
	\begin{tabular}{|c|c|c|}
		\hline
		\textbf{Author}                 & \textbf{Avg. RMSE} & \textbf{Avg. MAE} \\\hline
		D'Andrea                          & 745.9512              & 21.7650              \\\hline
		D'Andrea   (With Headwind)        & 919.2040              & 23.7520              \\\hline
		Dorling                           & 365.1678              & 18.3391              \\\hline
		Stolaroff                         & 291.6590              & 16.2421              \\\hline
		Kirchstein                        & 275.8223              & 13.8690              \\\hline
		Tseng                             & 131.3750              & 9.5263               \\\hline
		Proposed LSTM model & 36.2770                & 4.9080 \\ \hline               
	\end{tabular}
\end{table}

Among the five fundamental models, the D'Andrea model performs poorest with the highest averaged RMSE found for the model variant with the headwind being factored in. In contrast, the Tseng energy model achieves the best prediction result in that both performance metrics, i.e., averaged RMSE and averaged MAE, are lowest. This conclusion can also be further validated in Fig. \ref{fig15} where the Tseng model maintains a flat prediction throughout the duration of the evaluation which is clearly closest to the ground truth. However, none of these fundamental models achieves a promising result when compared with the proposed LSTM model where the averaged MAE is found as low as 4.908. In practice, this simply implies that on average the predicted energy only incurs 4.9080 watts bias compared with the realistic energy consumption. 

\subsection{Sensitivity Analysis}

\begin{figure}[htbp]
	\centering
	\includegraphics[width=\columnwidth, height=2.3in]{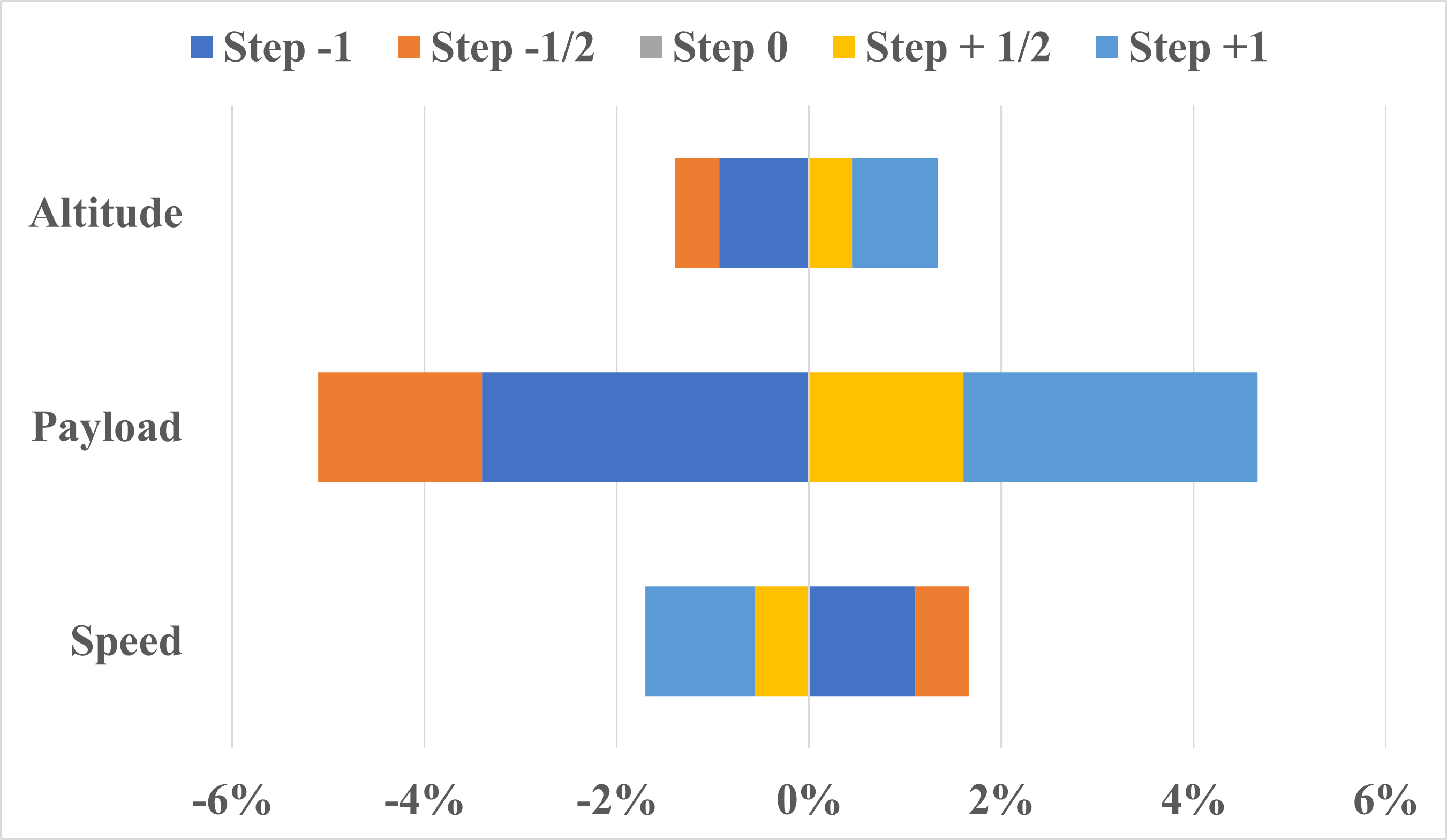}
	\caption{Sensitivity of power consumption to key features}\label{fig1}
\end{figure}

To further illustrate some interpretability of the trained LSTM model, we now carry out a sensitivity analysis for the learned model. In particular, we are interested in answering the following question, that is how the altitude, payload and speed as input features can contribute to the energy consumption for drones. Our results are shown in Table \ref{tabData} and Fig. \ref{fig1}. More specifically, the divergent bar chart in Fig. \ref{fig1} takes intervals at steps and half steps, where steps are taken as the minimum or maximum value of the input parameter, and half steps are the midpoints from the mean. Table \ref{tabData} shows the steps taken for each feature of interest. Fig. \ref{fig1} shows payload contributing the most towards change in overall power consumption followed by speed and altitude. Interestingly, the figure also demonstrates that the speed factor is negatively correlated to power consumption of the drone while the altitude factor can positively contribute to the power consumption. Such insight on speed is consistent with what we found in the Tseng model where airspeed is also negatively correlated with power consumption.

\begin{table}[htbp]
	\caption{Power Consumption Sensitivity Data Table}\label{tabData}
	\centering
	\begin{tabular}{|l|l|l|l|l|l|}
		\hline
		& \textbf{Step-1} & \textbf{Step-1/2} & \textbf{Step 0} & \textbf{Step+1/2} & \textbf{Step+1} \\ \hline
		Altitude (m)     & 25.00           & 43.75             & 62.50           & 81.25             & 100.00          \\ \hline
		Payload (g)      & 0               & 187.50            & 375.00          & 562.50            & 750.00          \\ \hline
		Speed (m/s)      & 4.00            & 6.00              & 8.00            & 10.00             & 12.00           \\ \hline
	\end{tabular}
\end{table}

\section{Conclusion} \label{conclusion}

In this paper, we propose a learning-based approach using the LSTM-based deep learning architecture for accurately modelling energy consumptions of drones. We have revealed the efficacy of the proposed model by comparing its performance with the five fundamental mathematical based energy models using a realistic dataset that can be publicly accessed in \cite{ref5}. Finally, we have also implemented a sensitivity analysis on the trained model with a view to provide some insights and explainability for the fine-tuned model. To conclude, we believe that the work presented in this paper is an important step towards using a data-driven method to understand energy consumption pattern for drones. However, we note that one obvious limitation in our current work is that the proposed deep learning model has not been validated comprehensively across different energy datasets. This challenge will be addressed as part of our future work. In addition, we shall investigate the effectiveness of using a federated learning based framework \cite{lstmSimulatedRef} to further improve accuracy of the proposed model.

\subsubsection{Acknowledgements} 

The author Carlos Muli would like to thank the ide3a team for kindly accommodating a scholarship which has made this research work possible. This work has also emanated from research supported in part by Science Foundation Ireland under Grant Number SFI/12/RC/2289\_P2, and the Entwine Research Centre at Dublin City University.

\bibliographystyle{ieeetran}
\bibliography{reference}

\begin{thebibliography}{10}
\providecommand{\url}[1]{#1}
\csname url@samestyle\endcsname
\providecommand{\newblock}{\relax}
\providecommand{\bibinfo}[2]{#2}
\providecommand{\BIBentrySTDinterwordspacing}{\spaceskip=0pt\relax}
\providecommand{\BIBentryALTinterwordstretchfactor}{4}
\providecommand{\BIBentryALTinterwordspacing}{\spaceskip=\fontdimen2\font plus
\BIBentryALTinterwordstretchfactor\fontdimen3\font minus
  \fontdimen4\font\relax}
\providecommand{\BIBforeignlanguage}[2]{{%
\expandafter\ifx\csname l@#1\endcsname\relax
\typeout{** WARNING: IEEEtran.bst: No hyphenation pattern has been}%
\typeout{** loaded for the language `#1'. Using the pattern for}%
\typeout{** the default language instead.}%
\else
\language=\csname l@#1\endcsname
\fi
#2}}
\providecommand{\BIBdecl}{\relax}
\BIBdecl

\bibitem{droneUse}
R.~Merkert and J.~Bushell, ``Managing the drone revolution: A systematic
  literature review into the current use of airborne drones and future
  strategic directions for their effective control,'' \emph{Journal of Air
  Transport Management}, vol.~89, p. 101929, 2020.

\bibitem{zaheer2016aerial}
Z.~Zaheer, A.~Usmani, E.~Khan, and M.~A. Qadeer, ``Aerial surveillance system
  using uav,'' in \emph{2016 thirteenth international conference on wireless
  and optical communications networks (WOCN)}.\hskip 1em plus 0.5em minus
  0.4em\relax IEEE, 2016, pp. 1--7.

\bibitem{thibbotuwawa2018energy}
A.~Thibbotuwawa, P.~Nielsen, B.~Zbigniew, and G.~Bocewicz, ``Energy consumption
  in unmanned aerial vehicles: A review of energy consumption models and their
  relation to the uav routing,'' in \emph{International Conference on
  Information Systems Architecture and Technology}.\hskip 1em plus 0.5em minus
  0.4em\relax Springer, 2018, pp. 173--184.

\bibitem{hu2020energy}
S.~Hu, Q.~Wu, and X.~Wang, ``Energy management and trajectory optimization for
  uav-enabled legitimate monitoring systems,'' \emph{IEEE Transactions on
  Wireless Communications}, vol.~20, no.~1, pp. 142--155, 2020.

\bibitem{ref1}
J.~Zhang, J.~F. Campbell, D.~C. Sweeney~II, and A.~C. Hupman, ``Energy
  consumption models for delivery drones: A comparison and assessment,''
  \emph{Transportation Research Part D: Transport and Environment}, vol.~90, p.
  102668, 2021.

\bibitem{model2}
K.~Dorling, J.~Heinrichs, G.~G. Messier, and S.~Magierowski, ``Vehicle routing
  problems for drone delivery,'' \emph{IEEE Transactions on Systems, Man, and
  Cybernetics: Systems}, vol.~47, no.~1, pp. 70--85, 2016.

\bibitem{ref4}
M.~A. Figliozzi, ``Lifecycle modeling and assessment of unmanned aerial
  vehicles (drones) co2e emissions,'' \emph{Transportation Research Part D:
  Transport and Environment}, vol.~57, pp. 251--261, 2017.

\bibitem{model3}
J.~K. Stolaroff, C.~Samaras, E.~R. O’Neill, A.~Lubers, A.~S. Mitchell, and
  D.~Ceperley, ``Energy use and life cycle greenhouse gas emissions of drones
  for commercial package delivery,'' \emph{Nature communications}, vol.~9,
  no.~1, pp. 1--13, 2018.

\bibitem{model4}
T.~Kirschstein, ``Comparison of energy demands of drone-based and ground-based
  parcel delivery services,'' \emph{Transportation Research Part D: Transport
  and Environment}, vol.~78, p. 102209, 2020.

\bibitem{model5}
C.-M. Tseng, C.-K. Chau, K.~M. Elbassioni, and M.~Khonji, ``Flight tour
  planning with recharging optimization for battery-operated autonomous
  drones,'' \emph{CoRR, abs/1703.10049}, 2017.

\bibitem{Prasetia}
A.~S. Prasetia, R.-J. Wai, Y.-L. Wen, and Y.-K. Wang, ``Mission-based energy
  consumption prediction of multirotor uav,'' \emph{IEEE Access}, vol.~7, pp.
  33\,055--33\,063, 2019.

\bibitem{ref5}
T.~A. Rodrigues, J.~Patrikar, A.~Choudhry, J.~Feldgoise, V.~Arcot, A.~Gahlaut,
  S.~Lau, B.~Moon, B.~Wagner, H.~S. Matthews \emph{et~al.}, ``In-flight
  positional and energy use data set of a dji matrice 100 quadcopter for small
  package delivery,'' \emph{Scientific Data}, vol.~8, no.~1, pp. 1--8, 2021.

\bibitem{ref2}
Z.~Fu, J.~Yu, G.~Xie, Y.~Chen, and Y.~Mao, ``A heuristic evolutionary algorithm
  of uav path planning,'' \emph{Wireless Communications and Mobile Computing},
  vol. 2018, 2018.

\bibitem{ref3}
J.~Modares, F.~Ghanei, N.~Mastronarde, and K.~Dantu, ``Ub-anc planner: Energy
  efficient coverage path planning with multiple drones,'' in \emph{2017 IEEE
  international conference on robotics and automation (ICRA)}.\hskip 1em plus
  0.5em minus 0.4em\relax IEEE, 2017, pp. 6182--6189.

\bibitem{additionalref1}
U.~C. {\c{C}}abuk, M.~Tosun, R.~H. Jacobsen, and O.~Dagdeviren, ``A holistic
  energy model for drones,'' in \emph{2020 28th Signal Processing and
  Communications Applications Conference (SIU)}.\hskip 1em plus 0.5em minus
  0.4em\relax IEEE, 2020, pp. 1--4.

\bibitem{model1}
R.~D'Andrea, ``Guest editorial can drones deliver?'' \emph{IEEE Transactions on
  Automation Science and Engineering}, vol.~11, no.~3, pp. 647--648, 2014.

\bibitem{lstmSimulatedRef}
M.~Liu, ``Fed-bev: A federated learning framework for modelling energy
  consumption of battery electric vehicles,'' in \emph{2021 IEEE 94th Vehicular
  Technology Conference (VTC2021-Fall)}.\hskip 1em plus 0.5em minus 0.4em\relax
  IEEE, 2021, pp. 1--7.

\end{thebibliography}

%
%
%
%
\end{document}